\documentclass[lettersize,journal]{IEEEtran}
\usepackage{amsmath,amsfonts}
\usepackage{amssymb}
\usepackage{algorithmic}
\usepackage{algorithm}
\usepackage{array}
\usepackage[caption=false,font=normalsize,labelfont=sf,textfont=sf]{subfig}
\usepackage{textcomp}
\usepackage{stfloats}
\usepackage{url}
\usepackage{verbatim}
\usepackage{graphicx}
\usepackage{cite}
\usepackage{bm}
\usepackage[table]{xcolor}
\begin{document}

\title{Comparative Evaluation of Deep-Learning based DoA Estimators for Real-Time ISAC Beamforming}
\author{Anjali K.S.$^{1}$, Manoj Kumar Panda$^{1}$, Chandramani Singh$^{2}$\\~\IEEEmembership{$^{1}$Department of Electronics and Communication Engineering,\\Amrita School of Engineering, Bengaluru, Amrita Vishwa Vidyapeetham, India\\ $^{2}$Department of Electronic Systems Engineering, Indian Institute of Science, Bangalore, India}}





\maketitle

\begin{abstract}
Integrated Sensing and Communication (ISAC) systems rely on accurate Direction-of-Arrival (DoA) estimation for effective beamforming and spectral-efficiency enhancement. While numerous classical and deep-learning-based DoA estimation methods have been reported in the literature, their suitability for real-time ISAC operation remains insufficiently understood because practical deployment requires consideration of both estimation accuracy and computational latency. This paper presents a unified evaluation framework for systematically comparing classical and learning-based DoA estimators within an ISAC beamforming pipeline. Representative classical and learning-based DoA estimators, together with an ensemble architecture, are integrated with Singular Value Decomposition (SVD)-based beamforming and evaluated under identical operating conditions. The impact of DoA estimation accuracy, inference latency, beamforming-weight generation time, and resulting effective spectral efficiency is investigated. Results show that different estimators occupy distinct operating points in the accuracy-latency-performance design space. Although learning-based approaches substantially reduce estimation time compared with MUSIC, the overall processing latency increasingly becomes dominated by the computational cost of SVD-based beamforming. Among the evaluated methods, the Regression Convolutional Neural Network (RCNN) exhibits a favorable balance between estimation accuracy, computational efficiency, and beamforming performance, making it a promising candidate for real-time ISAC deployment. The findings further suggest that future low-latency ISAC systems may benefit from beamforming strategies that reduce or avoid computationally intensive SVD-based beamforming-weight generation. 
\end{abstract}


\section{Introduction}
\IEEEPARstart{T}{he} rapid growth in wireless communication services and the rising demand for efficient spectrum utilization have motivated the integration of radar and communication functions on a common platform\cite{ref1}. This paradigm, widely studied under terms such as Joint Radar and Communication (JRC), Dual Function Radar Communication (DFRC), Integrated Sensing and Communication (ISAC), and Joint Sensing and Communication (JSAC), enables shared use of hardware and spectral resources. In ISAC systems, accurate Direction of Arrival (DoA) estimation plays a critical role, as it supports transmit beamforming for both reliable target detection and communication in the correct user direction. 

Traditional DoA estimation techniques, including MUSIC\cite{ref2}, Root‑MUSIC\cite{ref3}, ESPRIT\cite{ref4}, Fourier‑based spectral analysis\cite{ref5}, and MVDR\cite{ref6}, have been extensively applied in radar, wireless communication, remote sensing, and astronomy. However, these methods suffer from high execution time and degraded performance under low signal‑to‑noise ratio (SNR) conditions, motivating the adoption of deep learning (DL) approaches\cite{ref7}. Recent advances in DL have demonstrated significant improvements in DoA estimation accuracy and robustness. DL‑based methods can be broadly classified into classification‑based and regression‑based approaches. Classification methods discretize the angular space into predefined grids, achieving high accuracy but requiring large datasets, complex architectures, and longer training times\cite{ref8,ref9,ref10,ref11,ref12,ref13,ref14,ref15}. In contrast, regression‑based methods treat DoA as a continuous output, offering faster training, better generalization, and reduced computational cost, though they must be carefully designed to avoid instability\cite{ref16,ref17,ref18,ref19,ref20,ref21,ref22}. 

Even with these advancements, the evaluation metric used in most existing DoA estimation studies is root-mean-square error (RMSE) alone. A review on deep learning-based DoA estimation in \cite{ref_doa_review} reports  44\% of studies use RMSE alone as the evaluation metric. Even though RMSE provides estimation accuracy, in real-time system inference speed is equally important. For real-time ISAC operation, beamforming weights must be updated frequently to track target variations. In high-mobility scenarios, beam directions must be updated instantly to maintain alignment, and the resulting beam-training overhead and latency have been identified as critical bottlenecks limiting the practical deployment of predictive beamforming systems \cite{ref_beamforming_isac}. 

So, a systematic study is needed to evaluate DoA estimation accuracy-latency tradeoff in beamforming for real-time ISAC systems. While numerous studies focus on improving DoA estimation accuracy, relatively little attention has been paid to the impact of estimator complexity on real-time ISAC operation. Consequently, the most accurate estimator is not necessarily the most suitable estimator for real-time ISAC systems. Motivated by this gap, the present work evaluates existing DoA estimation and beamforming approaches not only in terms of estimation accuracy, but also prediction time and spectral efficiency, to provide a more complete picture of their suitability for ISAC deployment. 

In this study, DoA estimation based beamforming method for an ISAC transmitter base station is considered. The main contributions in this work are: 
\begin{itemize}
\item{An ISAC evaluation framework that integrates classical (MUSIC) and multiple deep-learning-based DoA estimation techniques with SVD-based beamforming is developed.}
\item{All these methods are evaluated under identical operating conditions with basic sine waves and Linear frequency modulated (LFM) signals for a broad range of SNR.}
\item{The DoA estimation accuracy and execution time of each method and its impact on beamforming weight generation is analyzed.}
\item{The effective spectral efficiency of downlink communication is checked to ensure communication performance in the specified ISAC scenario.}
\item{An ensemble model combining the regression models RDNN and RCNN, is also developed, which achieves the lowest RMSE among the evaluated estimators at low SNRs.}
\item{The comparative evaluation reveals that RCNN occupies a favorable operating point in the accuracy-latency-performance design space.} 
\end{itemize}

The rest of the paper is organized as section II with Literature Survey followed by System and Signal Models in section III. Section IV illustrates Methodology. In section V results are analyzed and in section VI insights are shared. The paper is concluded in section VII.

\section{Literature Survey}

Since 2017, DL based DoA estimation has gained significant attention due to its superior performance over traditional techniques \cite{ref7}. The accuracy of DoA estimation is primarily influenced by factors such as signal-to-noise ratio (SNR), number of sources, number of antenna elements, and number of snapshots \cite{ref7}. While CNN-based feature extraction has been widely applied in classification-oriented DoA estimation, its use in regression-based approaches remains less explored. Moreover, although raw sampled data can provide richer information than the covariance matrix—particularly in handling antenna array imperfections \cite{ref23}—the covariance matrix continues to be the preferred input owing to its computational efficiency. The DoA estimation methods and their limitations are summarized in Table~\ref{table:DoA_methods}

In beamforming, DoA is estimated first and based on that weights are predicted for beam forming in those directions. According to the literature reviewed, deep learning-based DoA estimation methods have demonstrated strong accuracy compared to classical approaches \cite{ref8,ref9,ref10,ref11,ref12,ref13,ref14,ref15,ref16,ref17,ref18,ref19,ref20,ref21,ref22}.The model from \cite{ref22}, which takes the full covariance matrix as input and uses convolutional layers followed by fully connected layers, is referred to as RCNN (Regression CNN) in this paper. In this work, we evaluate existing deep learning-based DoA estimation methods for the ISAC scenario described in \cite{ref1}.

Deep learning has been applied to ISAC systems to overcome the challenges of hard and unconstrained optimization by learning complex input–output mappings. For instance, an RIS-assisted ISAC system employing a CNN-LSTM architecture (CAP-Net) for DoA estimation from covariance matrices is proposed in \cite{ref24}, while HCL-Net in \cite{ref25} predicts beamforming matrices using historical channel information, assuming target DoAs. A transformer-based predictive beamforming scheme for vehicular ISAC networks was explored in \cite{ref26}, and a multi-subnet DL framework for DFRC beamforming was introduced in \cite{ref27}, where separate subnets learn communication-radar beamforming, and power allocation. The low-complexity and transferable framework in \cite{ref_lct} reduces beamforming optimization complexity through a three-module pipeline operating on Channel State Information (CSI). This relies on costly offline training and limit scalability for dataset generation and network training. Conventional ISAC beamforming relies on iterative non-convex optimization, such as successive convex approximation ((SCA), fractional programming, or their combination \cite{ref_fp_sca}, and the cost of each iteration is dominated by eigenvalue decomposition even in efficient solvers \cite{ref_mimo_nearness}. To meet the latency requirements of ISAC precoding, deep unfolding of SCA-based solvers has been proposed \cite{ref_dusca}, but it still retains a fixed-depth iterative structure. Reinforcement learning has also been applied to ISAC beamforming in dynamic settings such as UAV-enabled networks \cite{ref_uav_marl}, where beamforming is optimized jointly with trajectories over time.

Although these approaches demonstrate the potential of DL, they rely on complex architectures, do not address computational efficiency, and often assume DoA rather than estimating it directly. In contrast, in this study an ISAC evaluation platform is developed for comparing classical and deep-learning-based DoA estimators and the impact of DoA estimation errors  and complexity on downstream beamforming performance is quantified.

\begin{table}[tb]
\caption{DoA Estimation Methods and Their Limitations.}
\centering
\renewcommand{\arraystretch}{1.2}
\begin{tabular}{|>{\centering\arraybackslash}m{1.8cm}|>{\centering\arraybackslash}m{1.2cm}|>{\centering\arraybackslash}m{4.5cm}|}
\hline
\textbf{Methods} & \textbf{References} & \textbf{Limitations} \\
\hline
Traditional (e.g., MUSIC, ESPRIT) & \cite{ref2,ref3,ref4,ref5,ref6} & 
\begin{itemize}
\item High execution time.
\item Less accurate in noisy conditions.
\end{itemize} \\
\hline
Deep Learning (Classification) & \cite{ ref8, ref9, ref10, ref11, ref12, ref13, ref14, ref15} & 
\begin{itemize}
\item Limited to predefined angle grids
\item Inaccurate for off-grid DoAs.
\item High execution time and training time.
\item Requires large labeled datasets.
\end{itemize} \\
\hline
Deep Learning (Regression) & \cite{ref16,ref17,ref18,ref19,ref20,ref21,ref22} &
\begin{itemize}
\item Sensitive to training data distribution and outliers.
\item Vulnerable to unstable predictions in noisy conditions. 
\end{itemize} \\
\hline
\end{tabular}
\label{table:DoA_methods}
\end{table}

\section{System and Signal Models} 
\label{sec:models}

\subsection{ISAC Scenario}
\begin{figure}[t]
\centering
\includegraphics[width=2.5in]{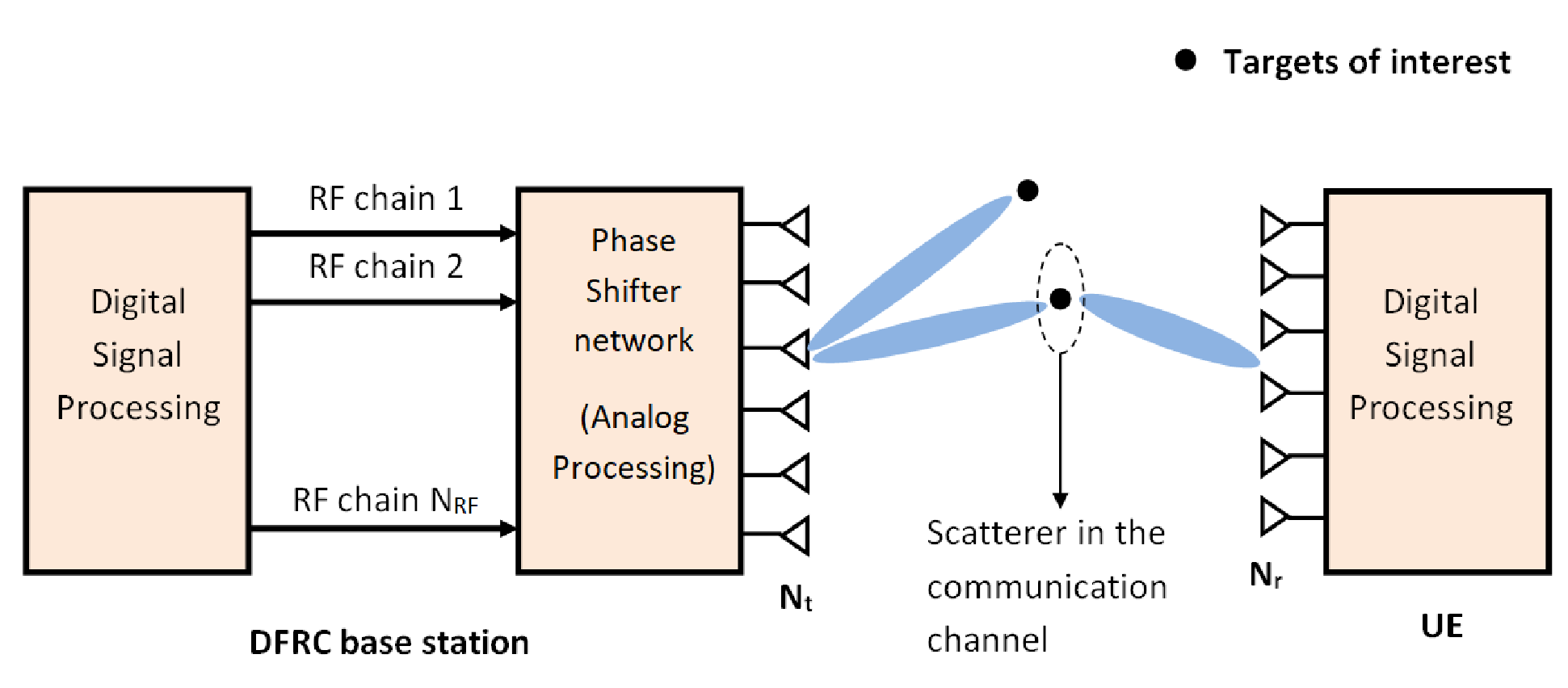}
\caption{Hybrid beamforming ISAC Scenario }
\label{fig_1}
\end{figure}

An ISAC base station (BS) which is used for both target detection and communication is considered here as in Fig. \ref{fig_1}. The base station operates in S-band and uses hybrid analog-digital (HAD) beamforming method mentioned in \cite{ref1}. We assume that the BS communicates with user equipment (UE) while detecting target via communication scatterer. Both BS and UE uses uniform linear arrays (ULA), where an $N_{t}$ antenna elements at BS and $N_{r}$ antenna elements at UE. The UE is assumed to have a completely digital beamforming structure.

In Fig. \ref{fig_1}, a sensing/communication environment is showed in which K number of targets/scatterers are dispersed randomly that are to be detected by the ISAC base station. 
We consider a uniform linear array (ULA), with consecutive elements separated by a distance $d$, at the base station. There are $K$ signal sources, which we call \textit{targets}\footnote{The $K$ signal sources can represent either $K$ communication transmitters or $K$ radar targets from which echoes are received in response to omnidirectional radar transmission. Eqn.~(\ref{eq1}) applies to both cases.}, with the direction of the $k$-th target given by an angle $\theta_k$, $k\,{=}\,1, \dots, K$. Each antenna element in the ULA receives a weighted sum of $K$ source signals, corresponding to the $K$ targets, with a wavelength $\lambda$. The source signals are either sinusoidal or linear frequency modulated (LFM). The signal received by every array element is sampled to produce $T$ samples or ``snapshots''. The (sampled) signal received by the ULA is given by 
\begin{equation}
\label{eq1}
\mathbf{y}(t)=\mathbf{A}(\theta) \mathbf{s}(t)+\mathbf{n}(t),  \, t\,{=}\,1, 2, \dots, T, 
\end{equation}
where $\mathbf{s}(t)\,{=}\,[s_{1}(t),s_{2}(t),....s_{K}(t)]$ denotes the source signal vector, with $s_{k}(t)\,{\in}\,\mathbb{C}$, $\forall t$, representing the complex-valued source signal pertaining to the $k$-th target, $\mathbf{n}(t)$ represents an additive white (complex) Gaussian noise signal, and $\mathbf{A}(\theta)=[\mathbf{a}(\theta_{1}),\mathbf{a}(\theta_{2}),..., \mathbf{a}(\theta_{K})] $ denotes the \textit{steering matrix} of dimension $N_{t} \times  K$, with the $k$-th steering vector, $\mathbf{a}(\theta_{k})$, given by 
\begin{equation}
\mathbf{a}(\theta_k) = 
\begin{bmatrix}
1, 
e^{j \frac{2\pi d}{\lambda} \sin(\theta_k)}, 
\dots ,
e^{j \frac{2\pi d}{\lambda} (N_t - 1) \sin(\theta_k)}
\end{bmatrix}^{T}.
\end{equation}

Even though every target reflects the echo wave back to the BS, not all target contributes to the communication channels that is spread between the BS and the UE. In mmWave channel modeling, the scattering model which usually have few scattering routes describes communication channels\cite{ref1}.  Out of K targets we assume only L are recognised in the communication channel, and $ L \leq Nr \leq Nt $. That is L data streams can be transmitted simultanously. For making the scenario simple, we assume BS is aware of both K and L, and BS has initially sent omnidirectional DL pilots (DP), and echo waves from K targets are approaching the ISAC BS. From the covariance matrix of the echo waves received at the base station, the DOA of all K targets $(\Theta)$, including the communication scatterers is estimated using DOA estimation methods. Additionally, the UE obtains the probing waveform through L scattering routes, which it uses to calculate L AoAs$(\Phi)$. 
After this, the BS will now be aware of $\Theta$  and the UE will have an estimate of $\Phi$ in a similar way. The BS then generates directional DL beams toward the DOA estimated by developing an ISAC beamformer, for precise observations and communication.

 To enable accurate data decoding at the UE, the joint beamformer design attempts to pre-equalize the communication channel effects here.

Representative deep learning-based and traditional approaches for DoA estimation are analyzed and compared in this paper. They are followed by DFRC beam former weight generation using SVD method mentioned in \cite{ref1}.

\section{Methodology} 
\label{sec:methodology}

The methodology adopted in the paper consists of:  Dataset Generation,  Architecture Design of Deep Learning (DL) Models, ISAC beamformer, Performance metrics for comparison.

\subsection{Dataset Generation} 

In this paper, we provide either the full covariance matrix of the received signal, $\mathbf{y}(t)$, or its upper triangular (UT) part as the input to the DL models. To that end, first, the source signal vector, $\mathbf{s}(t)$, is generated either as a vector of sinusoids or LFM signals, depending on the case under consideration. Next, the received signal, $\mathbf{y}(t)$, is calculated by applying (\ref{eq1}), where the noise samples, $\mathbf{n}(t)$, $\forall t$, and the angles $\theta_k$, $k\,{=}\,1, \dots, K$, that are required for the calculation are obtained as follows. 

In practice, the SNR may be unknown when a DL model is deployed after training. Hence, to ensure robustness against noise, we propose to train the DL models at multiple levels of SNRs. For generating each training instance, the SNR is randomly chosen from the levels $-13$ dB, $-10$ dB, $-5$ dB and $0$ dB. Once the SNR level is decided for a training instance, the corresponding noise variance, $\sigma^{2}_{N_{t}}$, is calculated, and the noise samples $\mathbf{n}(t)$, $\forall t$, are generated. The angles of the targets, $\theta_k$, $k\,{=}\,1, \dots, K$, that specify the DoAs of the source signals are calculated by adopting the method in \cite{ref20}. The possible DoAs are assumed to belong to the angular sector $[G_{\min},\,G_{\max}]$. The angular separation between the sources, $\Delta$, is taken from the set
$\left\{\Delta_{\min},\,\Delta_{\min}+ \Delta_d,\,\Delta_{\min}+2\Delta_d,\ldots, \Delta_{\max}\right\}$. Thus, the first DoA, $\theta_1$, is sampled at steps of  $\eta$ over the interval $[G_{\min},\,G_{\max} - (K - 1)\Delta]$, and the DoA of the $k^{th}$ source is $\theta_k\,{=}\,\theta_1 + (k - 1)\Delta$. This implies, $\theta_k\,{=}\,\theta_{k-1} +\Delta$. 

The covariance matrix of the signal, $\mathbf{y}(t)$, is given by 
\begin{equation}
\mathbf{R_{y}} = \mathbb{E} [\mathbf{y}(t)\mathbf{y}^{H}(t)] = \mathbf{A}(\theta)\mathbf{R}_{s}\mathbf{A}^{H}(\theta) + \sigma^{2}_{N_{t}}\mathbf{I}_{N_{t}}, 
\label{eq3}
\end{equation}
where $\mathbf{R_{s}}\,{=}\,\mathbb{E} [\mathbf{s}(t)\mathbf{s}^{H}(t)]$ is the covariance matrix pertaining to the source signals, $\mathbf{s}(t)$, and
$\mathbf{I}_{N_{t}}$ is an identity matrix of dimension $N_{t}\,{\times}\,N_{t}$. The matrix $\mathbf{R_{s}}$ is estimated as 
\begin{equation}
\mathbf{\widetilde{R}_s} = \frac{1}{T} \sum_{t=1}^{T} \mathbf{s}(t)\mathbf{s}^H(t), 
\label{eq4}
\end{equation}
and substituted into (\ref{eq3}) to calculate $\mathbf{\widetilde{R}_y}$. The instances of this estimate $\mathbf{\widetilde{R}_y}$ for various SNR levels, along with the ground truth angles $(\theta_{1},\dots,\theta_{k})$ are used as the training dataset. A different estimate $\mathbf{\widetilde{R}_y}'$, obtained as 
\begin{equation}
\mathbf{\widetilde{R}_y}' = \frac{1}{T} \sum_{t=1}^{T} \mathbf{y}(t)\mathbf{y}^H(t), 
\label{eq5}
\end{equation}
is utilized for testing to reflect the fact that when the DL model is deployed after training, it could only use the samples of $\mathbf{y}(t)$ to estimate the DoAs. 
\subsection{Architecture Design of Deep Learning (DL) Models}
The architecture of the estimation models are same as provided in their references. The input type, size and layer type is provided in Table.~\ref{table:model_summary}. All the models generally use covariance matrix as input, in which some models use only upper triangular(UT) elements of covariance matrix. As the matrix contains complex numbers, mostly these elements are fed by externally splitting the real and imaginary part, where in RCNN it is left as is to be internally handled by the CNN and DNN layers. 
\begin{table}[tb]
\caption{ Model Architectures and Configurations.}
\centering
\renewcommand{\arraystretch}{1.2}
\begin{tabular}{|p{1.0cm}|p{2.0cm}|p{1.22cm}|p{1.8 cm}|}
\hline
\textbf{Model Name} & \textbf{Covariance matrix input} 
& \textbf{Input Size} & \textbf{Layer type}\\
\hline
CLASS\cite{ref8} &  Full matrix (real,
imaginary, phase) 
& $16\times16\times3 $ & 4 Convolutional + FC\\
\hline
RDNN\cite{ref20} &  UT  elements (Real and imaginary) 
& $1 \times 240$ & FC only \\
\hline
LDNN \cite{ref21} & Average of UT
elements along super-diagonals (Real and imaginary) 
& $1 \times 30$& FC only  \\ 
\hline
RCNN \cite{ref22} & Full Covariance matrix 
 & $16 \times 16$& 1 Convolutional +FC  \\
\hline
 
\end{tabular}
\label{table:model_summary}
\end{table}

\subsection{ISAC beamformer} 

Here we use DFRC beamformer framework mentioned in \cite{ref1}, to design hybrid beamforming weights $\mathbf{F}_{\text{RF}}$ and $\mathbf{F}_{\text{BB}}$ which are analog and digital beamforming weights respectively.

The rank  $L$ communication channel matrix in the ISAC scenario can be expressed as:

\begin{equation}
    \mathbf{H} = \mathbf{B}(\bm{\Phi}) \operatorname{diag}(\bm{\beta}) \mathbf{A}^T (\bm{\Theta}_1).
\end{equation}
Each notation in the equation (6) are as follows.
$\mathbf{B}(\bm{\Phi})$ represents the steering matrix of the UE’s antenna array with steering vectors $\mathbf{b}(\phi_l)$ in the AoA directions $ \phi $. That is,
\begin{equation}
    \mathbf{B}(\bm{\Phi}) = \big[ \mathbf{b}(\phi_1), \dots, \mathbf{b}(\phi_L) \big],
\end{equation}
\begin{equation}
    \bm{\Phi} = \big[ \phi_1, \dots, \phi_L \big],
\end{equation}
When considered from UE, the DL AoDs are part of the set $\bm{\Theta} = \big[ \theta_1, \dots, \theta_K \big]$ of radar targets that are visible from the BS. 
\begin{equation}
   \bm{\Theta}_1= \{ \theta_1, \dots, \theta_L \} \subseteq \bm{\Theta}.
\end{equation}

The $\mathbf{A}(\bm{\Theta}_1)$ represents the steering matrix with steering vectors $\mathbf{a}(\theta_l)$, which are DL AoD of the $l$ th scattering path,
\begin{equation}
    \mathbf{A}(\bm{\Theta}_1) = \big[ \mathbf{a}(\theta_1), \dots, \mathbf{a}(\theta_L) \big],
\end{equation}
And $\bm{\beta}$ denote the complex scattering coefficient matrix
\begin{equation}
    \bm{\beta} = \big[ \beta_1, \dots, \beta_L \big]^T,
\end{equation}

where $\mathbf{\beta_l}$ represents the complex scattering coefficient of the $l$ th scattering path which can be found by  APES algorithm as mentioned in \cite{ref1} that follow gaussian distribution. $\mathbf{a}(\theta_l)$ and $\mathbf{b}(\phi_l)$ can be derived from equation (2) in a similar way.

In order to reduce interference and guarantee correct signal reception at the receiver, inverting the channel matrix (or computing its pseudo-inverse) is necessary for equalization in beamforming. As neither $\mathbf{H} \mathbf{H}^H$ nor $\mathbf{H}^H \mathbf{H}$ is invertible, a pseudo-inverse of $\mathbf{H}$ is unobtainable. So both transmit and receive beamformings are necessary for equalizing the channel.
When $\tilde{H} $ can be defined as 
\begin{equation}
\mathbf{\tilde{H}}\triangleq \text{diag}(\boldsymbol{\beta}) \mathbf{A}^T(\boldsymbol{\Theta}_1) 
\end{equation}
it is possible to express channel matrix as
\begin{equation}
    \mathbf{H} = \mathbf{B} (\bm{\Phi}) \tilde{\mathbf{H}}.
\end{equation}
Both $\tilde{\mathbf{H}}$ estimated at BS and $\mathbf{B} (\bm{\Phi})$ estimated at UE will have a full rank of  $L$. The corresponding zero-forcing (ZF) beamformer at base station can be formulated as:

\begin{equation}
    \mathbf{F}_{\text{BS}} = \tilde{\mathbf{H}}^H \left( \tilde{\mathbf{H}} \tilde{\mathbf{H}}^H \right)^{-1},
\end{equation}
And corresponding zero-forcing (ZF) beamformer at UE can be formulated as
\begin{equation}
    \mathbf{W}_{\text{UE}} = \left( \mathbf{B}^H (\bm{\Phi}) \mathbf{B} (\bm{\Phi}) \right)^{-1} \mathbf{B}^H (\bm{\Phi}).
\end{equation}

Here the hybrid array at the BS can only roughly achieve $\mathbf{F}_{\text{BS}}$, and the $\mathbf{W}_{\text{UE}}$  at UE is a fully-digital beamformer. The beamformer $\mathbf{F}_{\text{D}} = \mathbf{F}_{\text{RF}} \mathbf{F}_{\text{BB}}$ should be designed in such a way it steer beams towards all $K$ targets.
This is corresponding to designing the covariance matrix of the transmit signal, which is formulated as:
\begin{equation}
    \mathbf{R}_s = \mathbb{E} \left[ \mathbf{F}_{\text{D}} \mathbf{s} \mathbf{s}^H \mathbf{F}_{\text{D}}^H \right] 
    = \mathbf{F}_{\text{D}} \mathbb{E} \left[ \mathbf{s} \mathbf{s}^H \right] \mathbf{F}_{\text{D}}^H
    = \mathbf{F}_{\text{RF}} \mathbf{F}_{\text{BB}} \mathbf{F}_{\text{BB}}^H \mathbf{F}_{\text{RF}}^H.
\end{equation}

A simple approach to construct the  $\mathbf{F}_{\text{RF}}$ is to  formulate each column of $\mathbf{F}_{\text{RF}}$ with  steering vector  of corresponding K  angles that gives:

\begin{equation}
    \mathbf{F}_{\text{RF}} (:, i) = \mathbf{a}^* (\theta_i), \quad \forall i=1,2..K.
\end{equation} 

Now $F_{BB}$ can be obtained by a global optimal closed form solution for a non-convex optimization problem defined in \cite{ref1} as, 

\begin{equation}
    \mathbf{F}_{\text{BB}} = \sqrt{\frac{P_T}{K N_t}} \tilde{\mathbf{U}} \tilde{\mathbf{V}}^H,
\tag{17}\end{equation}

where $\tilde{\mathbf{U}} \tilde{\bm{\Sigma}} \tilde{\mathbf{V}}^H$  can be solved using  singular value decomposition (SVD) method as:

\begin{equation*} {\mathbf {\tilde U{\tilde \Sigma } }}{{\mathbf {\tilde V}}^{H}} = {\mathbf {F}}_{RF}^{H}\left [{ {{{\mathbf {F}}_{BS}},{{\mathbf {F}}_{aux}}} }\right]\tag{18}\end{equation*}

Here $\mathbf{F}_{\text{BS}} $ is defined as in  (14), and $\mathbf{F}_{\text{aux}} \in \mathbb{C}^{N_t \times (K-L)}$ is an auxiliary matrix which is considered as a  null-space projection (NSP) matrix, such that $\tilde{H} \mathbf{F}_{\text{aux}} = 0 $ .

\subsection{Spectral Efficiency Computation}

The performance of the DL communication is evaluated by computing the  effective spectral efficiency (ESE). The signal vector received by UE can  be written as:

\begin{align*} {{\mathbf {y}}_{DL}}=&{\mathbf {B}}\left ({\Phi }\right)\mathrm {diag} \left ({{\boldsymbol \beta } }\right){{\mathbf {A}}^{T}}\left ({{\Theta _{1}} }\right){{\mathbf {F}}_{RF}}{{\mathbf {F}}_{BB}}{\mathbf {s}} + {{\mathbf {n}}_{DL}} \\\triangleq&{\mathbf {H}}{{\mathbf {F}}_{RF}}{{\mathbf {F}}_{BB}}{\mathbf {s}} + {{\mathbf {n}}_{DL}},\tag{19}\end{align*}

Provided $\mathbf{n}_{DL}$ is the AWGN noise added and $\mathbf{s}=[s1;s2]$  signals transmitted.
Here only  $\mathbf{s}_1$ is used for DL communication, whereas both  $\mathbf{s}_1$ and $\mathbf{s}_2$ are used for radar target detection. $\mathbf{s}_2$ includes no usable information because the communication channel only allows the transmission of $L$ independent data streams.
When the received signal vector is multiplied by $ W_{UE}$ and $\sqrt{\rho_{DL}}$ , the post processing signal received at the UE is:
\begin{equation}
    \tilde{\mathbf{y}}_{DL} =
    \sqrt{\rho_{DL}} \mathbf{W}_{UE} \mathbf{H} \mathbf{F}_{RF} \mathbf{F}_{BB} \mathbf{s} + \mathbf{W}_{UE} \mathbf{n}_{DL}.
    \label{ydl}
\tag{20}\end{equation} 
where, ${\rho_{DL}}$ is the average power received at UE.
The designed beamforming matrix can be splitted as :
\begin{equation*} {{\mathbf {F}}_{D}} = {{\mathbf {F}}_{RF}}{{\mathbf {F}}_{BB}} = {{\mathbf {F}}_{RF}}\left [{ {{{\mathbf {F}}_{BB,1}},{{\mathbf {F}}_{BB,2}}} }\right],\tag{21}\end{equation*} 
where \( \mathbf{F}_{BB,1} \in \mathbb{C}^{K \times L} \)   and  \( \mathbf{F}_{BB,2} \in \mathbb{C}^{K \times (K-L)} \).
Now equation ~\eqref{ydl} can be written as:
\begin{align*} {{{\mathbf {\tilde y}}}_{DL}}=&\underbrace {\sqrt {\rho _{DL}} {{\mathbf {W}}_{UE}}{\mathbf {H}}{{\mathbf {F}}_{RF}}{{\mathbf {F}}_{BB,1}}{{\mathbf {s}}_{1}}}_{\text {Useful Signal}} \\&+ \underbrace {\sqrt {\rho _{DL}} {{\mathbf {W}}_{UE}}{\mathbf {H}}{{\mathbf {F}}_{RF}}{{\mathbf {F}}_{BB,2}}{{\mathbf {s}}_{2}}}_{\text {Interference}} \\&+\, {{\mathbf {W}}_{UE}}{\mathbf {n}}_{DL}.\label{ydl_split}\tag{22}\end{align*}
The second term in ~\eqref{ydl_split} represents interference at the UE, as it contains no useful information.
The spectral efficiency is computed as:
\begin{align*} \mathbf{R}_{DL} = \log \det \left ({\!\begin{gathered} {{\mathbf {I}}_{L}} + \frac {\rho _{DL} }{L}{\mathbf {R}}_{in}^{ - 1}{{\mathbf {W}}_{UE}}{\mathbf {H}}{{\mathbf {F}}_{RF}}{{\mathbf {F}}_{BB,1}} \\ \;\;\;\;\;\;\; \times {\mathbf {F}}_{BB,1}^{H}{\mathbf {F}}_{RF}^{H}{{\mathbf {H}}^{H}}{\mathbf {W}}_{UE}^{H} \\ \end{gathered} \!}\right)\!,\tag{23} \label{SE_eq}\end{align*} 
where \( \mathbf{R}_{in} \) represents the covariance matrix of the interference plus noise, given by:
\begin{align*}&\hspace {-.5pc}{{\mathbf {R}}_{in}} = \rho _{DL} {{\mathbf {W}}_{UE}}{\mathbf {H}}{{\mathbf {F}}_{RF}}{{\mathbf {F}}_{BB,2}}{\mathbf {F}}_{BB,2}^{H}{\mathbf {F}}_{RF}^{H}{{\mathbf {H}}^{H}}{\mathbf {W}}_{UE}^{H} \\& \qquad \qquad \qquad \qquad\displaystyle {+\, \sigma _{c}^{2}{{\mathbf {W}}_{UE}}{\mathbf {W}}_{UE}^{H}. } \tag{24}\end{align*}

The effective spectral efficiency for highly mobile scenarios is \cite{ref_ESE}:
\begin{align*} R_{\mathrm {eff}}=&~\left ({1-\frac {T_{\mathrm {tr}}}{T_{\mathrm {B}}}}\right) \times {R_{DL}} \tag{25}\end{align*} 
where $T_{B}$ is the channel beam coherence time  which represents the average time over which the beams stay aligned and first $T_{tr}$ time of every beam coherence time is allocated for the channel training/beamforming design . 

\subsection{Estimation Accuracy and Computational Time}
The DoA estimation accuracy is evaluated by the root mean square error (RMSE) at different SNR levels, given by 
\begin{equation}
\mathrm{RMSE} \;=\;
\sqrt{\frac{1}{NK} 
      \sum_{n=1}^{N}
      \sum_{k=1}^{K}
      \bigl(\hat{\theta}_{n,k}-\theta_{n,k}\bigr)^{2}},
      \tag{25}
\label{eq:rmse}
\end{equation}
where $N$ is the number of trials during testing experiments. The term \(\hat{\theta}_{n,k}\) denotes the estimated DoA of the \(k\)-th source signal in the \(n\)-th trial, while \(\theta_{n,k}\) represents the corresponding true DoA. 

For evaluating the execution time, the prediction time of the DL models including the pre-processing time is obtained. Although asymptotic complexity analysis is commonly used for algorithmic comparison, the considered ISAC scenario involves fixed and relatively small problem dimensions ($N_t\,{=}\,16$, $K\,{=}\,L\,{=}\,2$). Under such conditions, asymptotic expressions may not accurately reflect practical implementation latency. Therefore, measured execution time, including pre-processing overhead, is adopted as the primary computational-cost metric in this work. 

\section{Results and Discussions} 

In this section simulation settings for  each methods and their simulation results are presented to analyze the accuracy, execution time, and spectral efficiency to find which method demonstrate higher accuracy-latency tradeoff. Each method \cite{ref1,ref8,ref20,ref21,ref22} is examined for two cases when transmitted signals  are basic sine waves and  LFM signals. All the methods are implemented using MATLAB in Windows operating system running on the 12th Gen Intel(R) Core(TM) i9-12900H processor  at 2.5 GHz and with an  NVIDIA GeForce RTX 3080 Ti Laptop GPU.

\subsection{Simulation Settings}
We simulated the scenario with $K\,{=}\,L\,{=}2 $ targets, generating signals of frequency $f\,{=}\,2.4$ GHz, and a ULA of $N_{t}\,{=}\,16$ elements uniformly separated by a half-wavelength, i.e., $d\,{=}\,\lambda/2$, where $\lambda\,{=}\,c/f$, and $c$ is the speed of light in vacuum. Signals of 1 ms duration are sampled to get $T\,{=}\,1000$ snapshots. For LFM signals \cite{b34}, a bandwidth of 100 kHz is considered.  We assume target of interest at  direction $\theta_1=-10.11^\circ$  and $\theta_2=26.11^\circ$. To generate the training dataset, we set $G_{min}\,{=}\,-60^{\circ}$, $G_{max}\,{=}\,60^{\circ}$, $\Delta_{min}\,{=}\,2^{\circ}$, $\Delta_{max}\,{=}\,60^{\circ}$, $\Delta_{d}\,{=}\,2^{\circ}$, and $\eta\,{=}\,1^{\circ}$.

All the DL models are trained using multiple SNR levels of $\{-13,-10,-5,0\}$~dB with corresponding true DoA angles, resulting in 10,680 training instances. The networks employs  `Adam' optimizer with an initial learning rate of 0.001, reduced by a factor of 0.1 every 10 epochs up to 200 epochs. A mini-batch size of 32 is used. To mitigate overfitting, L2 regularization ($\lambda=0.001$) and early stopping (patience=10) are applied. Training data are shuffled at initialization and after each epoch, while gradient clipping (threshold=1) stabilizes updates. Model performance is monitored on a validation set, evaluated every 100 iterations.

\subsection{DoA Estimation Accuracy}
\begin{figure}[!t]
\centering
\includegraphics[height=5.5cm,width=3.0in]{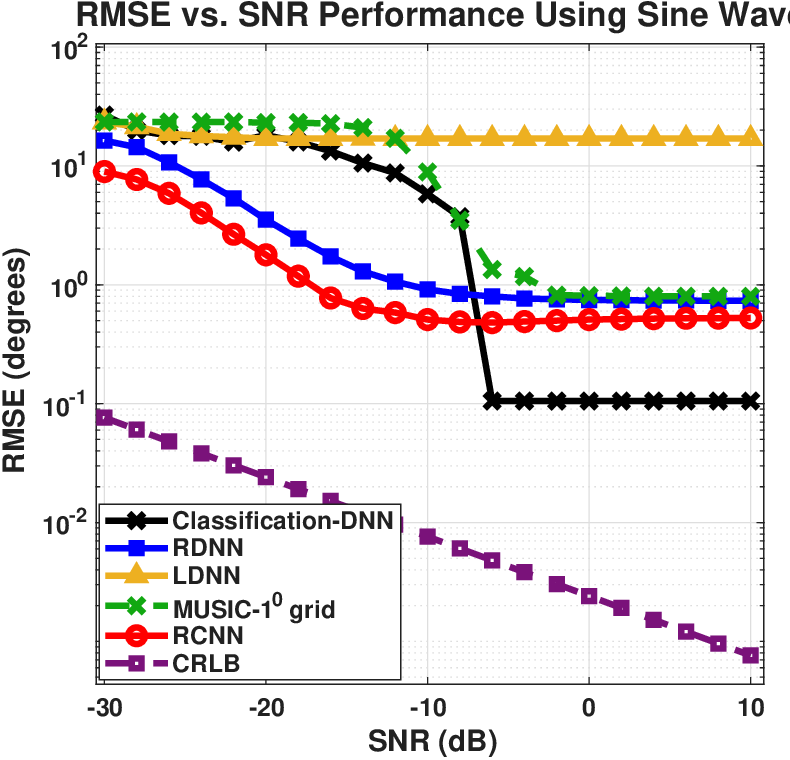}
\caption{RMSE vs. SNR performance with sinusoidal signals.}
\label{result1}
\end{figure}

\begin{figure}[!t]
\centering
\includegraphics[height=5.5cm,width=3.0in]{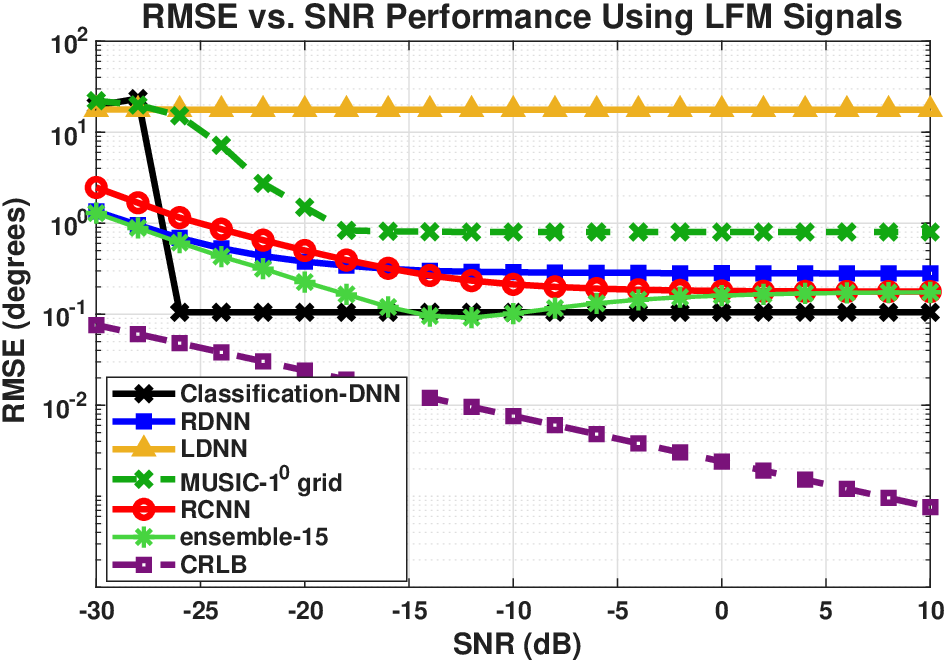} 
\caption{RMSE vs. SNR performance with LFM signals.}
\label{result2}
\end{figure}

The RMSE vs. SNR performance of the DoA estimation methods with sinusoidal and LFM signals are depicted in Figs.~\ref{result1} and~\ref{result2}, respectively, for SNRs from $-30$ dB to $10$ dB. The associated Cramer-Rao Lower Bounds (CRLB) is obtained from \cite{ref25}. 

From Figs.~\ref{result1} and~\ref{result2} observations made are as follows:
\begin{itemize}
\item [O1.] The performance with LFM signals is, in general, better and less sensitive to the SNR level than that with sinusoidal signals because of the inbuilt noise immunity of the LFM signals due to the chirp waveform. 
\item[ O2.] Classification method shows better accuracy at higher SNRs for both LFM and sinusoidal cases.
\item [O3.] For sinusoidal signals at higher SNR range the  RCNN method provides second-highest accuracy after the classification method. At lower SNRs, however, RCNN surpasses the classification method in accuracy.
\item [O4.] LDNN exhibited significant sign ambiguities in angle prediction, resulting in substantially higher RMSE for both sinusoidal and LFM signals. 
\item [O5.] For LFM signals, from -15dB to higher SNRs the  RCNN shows second highest accuracy. At very low SNRs RDNN shows better accuracy. The inbuilt noise immunity of LFM signal together with the usage of less noise upper triangular elements of covariance matrix  as input to the network might be the reason for this.
\item[O6.] For LFM signals, as RCNN is better from -15 dB to higher SNRs and RDNN is better at lower SNRs, an ensemble model is tried and it works better than both models. At  SNR range from -14dB to -10dB, ensemble model provides lower RMSE than that exhibited by the classification method as well.

For ensemble method, the prediction is done by using weighted sum of prediction of both models. That is prediction of ensemble model, $P=(w1 \times P1) +(w2\times P2) $ where $w1$ and $w2$ are weights and $P1$ and $P2$  are  prediction from RDNN and RCNN respectively. We used a soft switching using sigmoid function for weight generation according to SNR for calculating ensemble  prediction. That is,  $w_1 = \frac{1}{1 + e^{0.2(\mathrm{SNR} + 15)}},$  where the factors  0.2 controls the smoothness and 15 adjusts the transition region centered at -15dB and  $w_2 = 1 - w_1$. When SNR=-15dB, $w1=w2=0.5$, after -15dB SNR to higher SNRs w1 decays and w2 will be higher according to the equation. So switching of models happens from lower to higher SNRs accordingly. 
\item[O7.] The ensemble model shows the maximum  improvement in RMSE of $1.1770^\circ$ at -30dB SNR and minimum improvement in RMSE of $0.0028^\circ$ at 10dB SNR.

\end{itemize}

Apart from the above proposed method for ensemble model, another binary classifier DNN model can also be developed to identify  which model prediction has to be considered at a particular SNR range based on output of both the models.

\subsection{Radiation pattern comparison}
\begin{figure}[!t]
\centering
\includegraphics[height=5.5cm,width=3.0in]{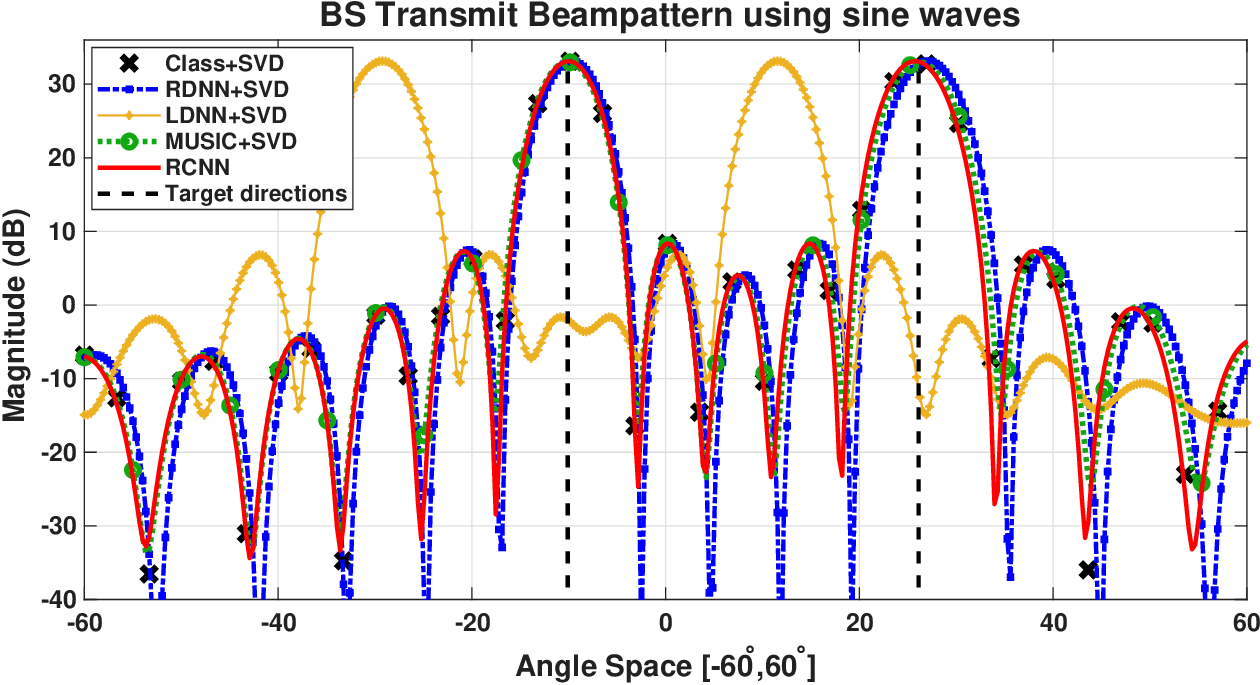}
\caption{Transmitter beam pattern for BS transmitting sine waves}
\label{BEAM_SINE}
\end{figure}

\begin{figure}[!t]
\centering
\includegraphics[height=5.5cm,width=3.0in]{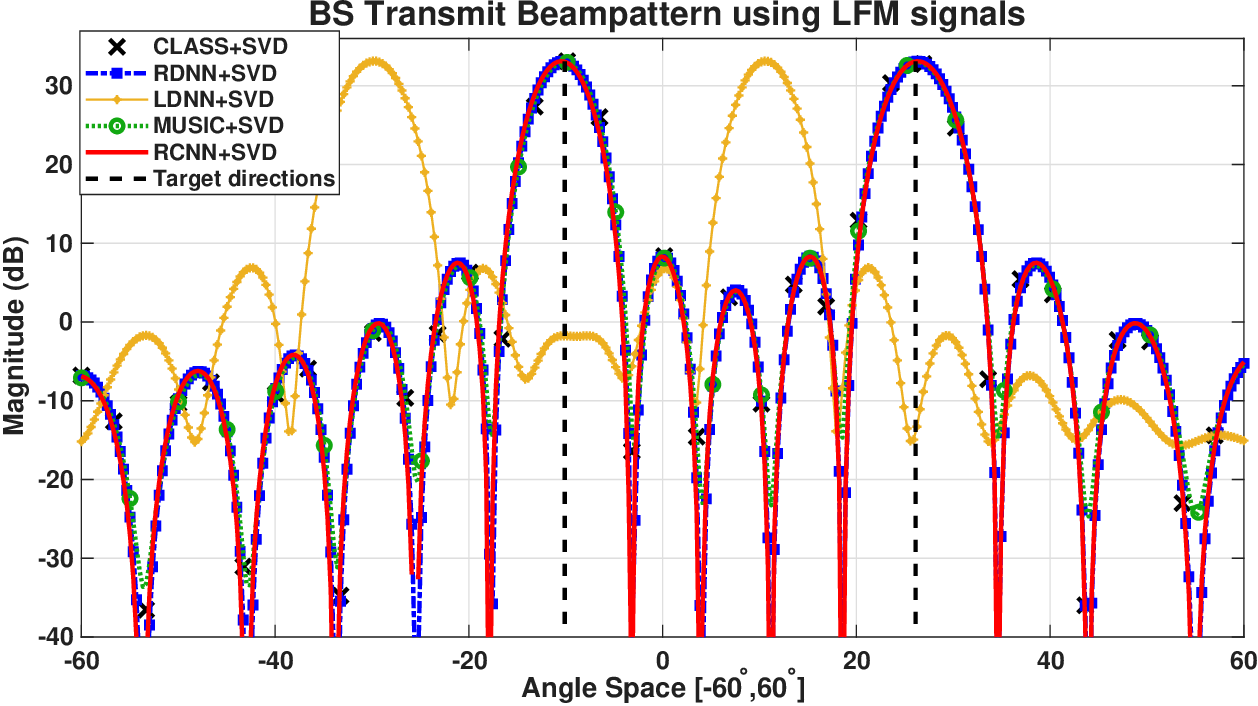}
\caption{Transmitter beam pattern for BS transmitting LFM signals.}
\label{BEAM_LFM}
\end{figure}

Based on the angles estimated using different methods beam
forming weights are derived using the SVD method. The beampatterns obtained from those weights are plotted for both the cases (LFM and sine wave) in Fig.~\ref {BEAM_SINE} and Fig.~\ref {BEAM_LFM} at 10 dB SNR.
As LDNN performance was not accurate as other methods, the angles predicted by this method were also not accurate.
Hence in the beampattern, mainlobes are in different directions for LDNN method. All other methods has obtained mainlobes in the necessary directions itself. 

\subsection{Computational Time Analysis}
\begin{figure}[!t]
\centering
\includegraphics[height=4.5cm,width=3.0in]{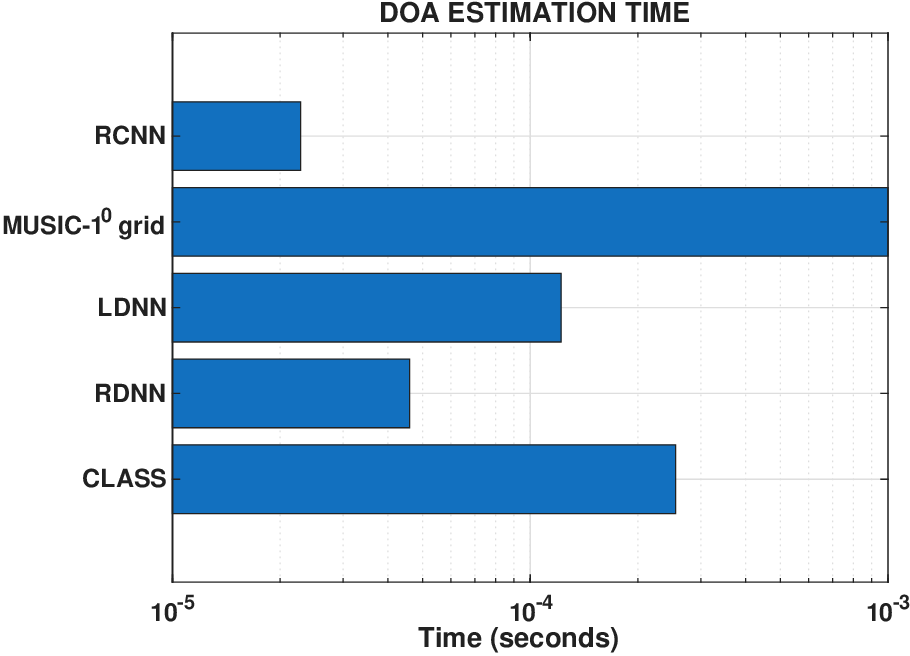}
\caption{Prediction time needed for each method of DoA estimation}
\label{DOA_TIME}
\end{figure}

The prediction time comparison for all the DL models and MUSIC algorithm, to estimate DoA including  the pre-processing time, is provided in Fig. ~\ref {DOA_TIME}. As all the method requires covariance matrix generation, pre-processing time is taken after covariance matrix generation process. The extraction of upper triangular elements and external splitting of complex numbers require more pre-processing time compared to feeding the network with the full covariance matrix and utilizing internal complex number handling. It can be inferred from  Fig.~\ref {DOA_TIME} that the RCNN incurs the least computational time compared to other  DL models, and traditional method.

\begin{figure}[!t]
\centering
\includegraphics[height=4.5cm,width=3.0in]{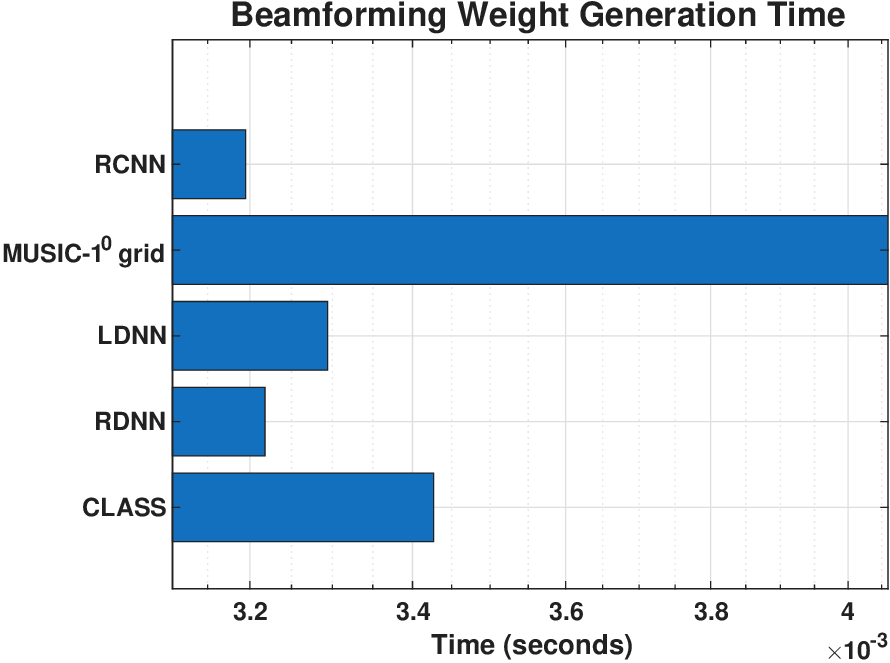}
\caption{Beamforming weight generation time for each method}
\label{BEAM_time}
\end{figure}

Fig.~\ref {BEAM_time} shows the time for generating beamforming weights after DoA estimation using each method. The SVD beamforming process takes around 0.0032s. Here it can be inferred very clearly that, as the DoA estimation time for RCNN is less, it achieves less beamforming weight generation time also. As this method offers the most favorable accuracy-ESE tradeoff, it points to the  efficiency  of RCNN compared to all other methods here. In ensemble method RDNN and RCNN will be working parallel and from the predictions of both models ensemble model prediction is finally made.
So the DoA prediction time for ensemble model will be  RDNN's (as that is higher than RCNN)  DoA prediction time and time to run the steps to find the ensemble prediction. 

\begin{figure}[!t]
\centering
\includegraphics[height=5.5cm,width=3.0in]{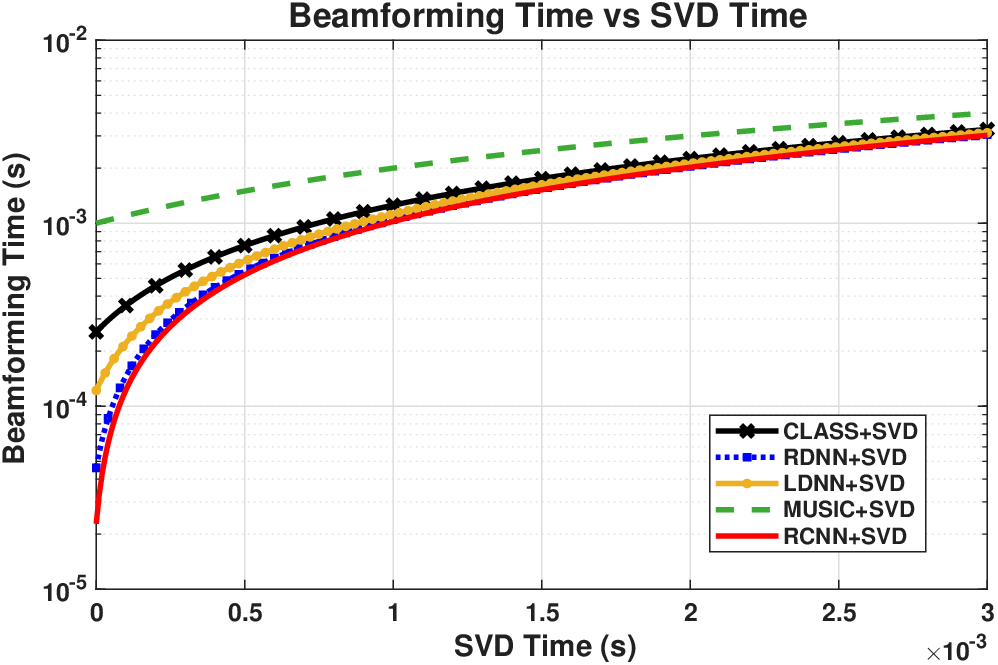}
\caption{Beamforming weight generation time Vs. SVD time analysis}
\label{BEAM_vs_svd}
\end{figure}

In Fig.~\ref {BEAM_vs_svd} beamforming time with increase in SVD time is analysed. At the starting of the graph, when SVD time is zero, it shows basically the DoA estimation time of each method. We could see that although RCNN substantially reduces DoA estimation latency, the overall beamforming latency remains increasingly dominated by the SVD stage. Consequently, the benefits of faster DoA estimation cannot be fully exploited unless the computational burden of beamforming-weight generation is also reduced. 

 \subsection{Spectral Efficiency Analysis}
\begin{figure}[!t]
\centering
\includegraphics[height=5.25cm,width=3.0in]{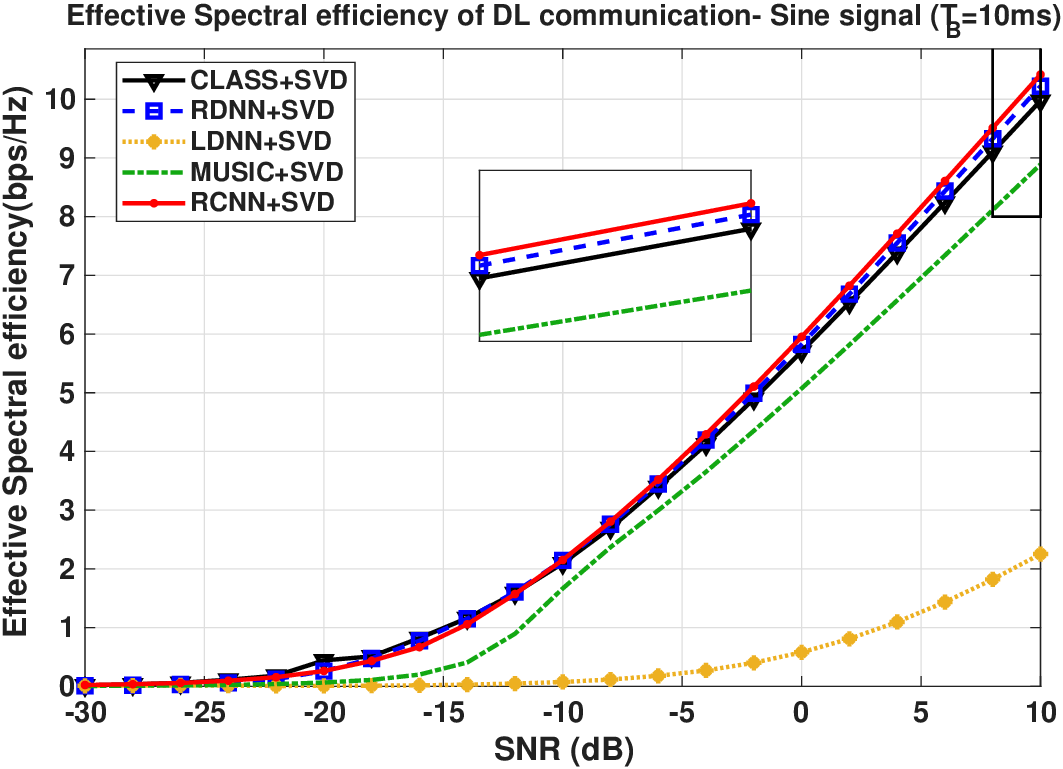}
\caption{Spectral efficiency for DL communication for sine wave}
\label{SE}
\end{figure}

\begin{figure}[!t]
\centering
\includegraphics[height=5.25cm,width=3.0in]{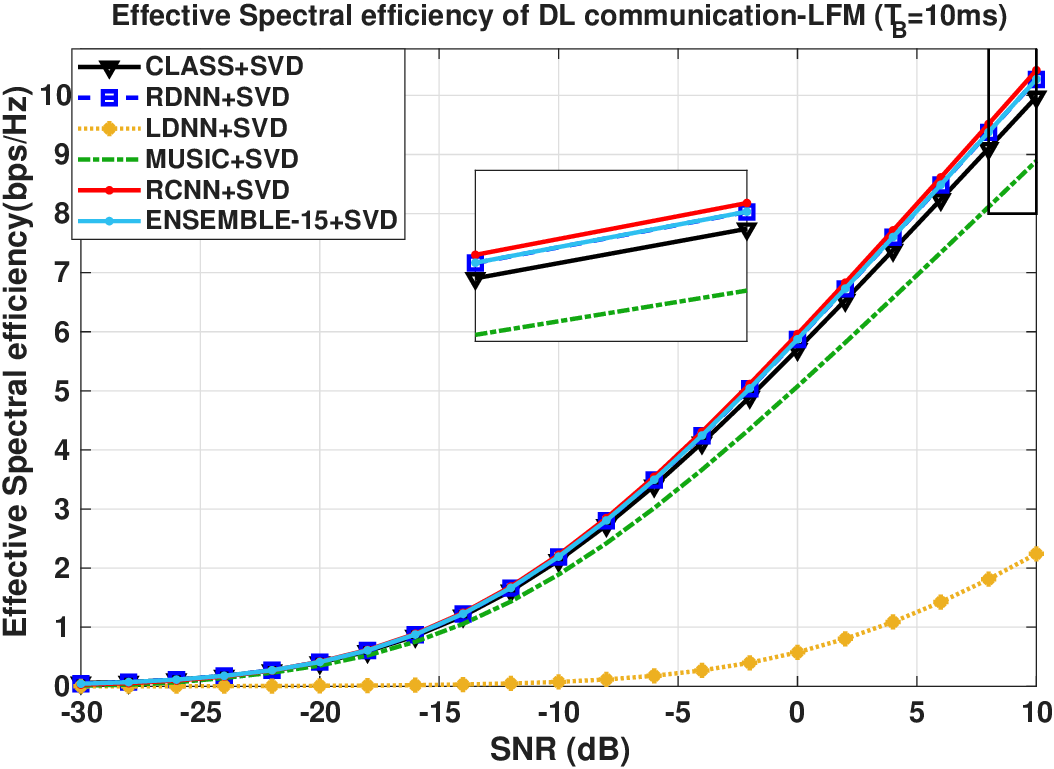}
\caption{Spectral efficiency for DL communication for LFM signal}
\label{SE}
\end{figure}

Fig.~\ref {SE} shows the SNR vs Effective spectral efficiency curve  of DL communication computed by equation (25) for $T_{B}=10$ms. As SNR increases, spectral efficiency is increasing for all the methods considered. For both sine and LFM signals the RCNN+SVD method provides higher spectral efficiency. As LDNN's beam was not in the correct direction, the effective spectral efficiency of LDNN+SVD is very less compared to all other methods. The ensemble method shows the almost the same spectral efficiency as RDNN method throughout the SNR range considered.

\begin{table}[t]
\caption{Performance comparison at 10 dB using LFM signals. 
\label{Performance comparison}}
\centering
\begin{tabular}{|c||c|c|c|c|c|}
\hline
Methods & DoA RMSE & DoA Pred.Time(s)  & ESE (bps/Hz)\\
\hline
CLASS & \bf{0.1051} & $2.55 \times 10^{-04}$  & 9.9738\\
\hline
RDNN & 0.2809 & $ 4.6 \times 10^{-5} $  & 10.2694\\
\hline
LDNN & 17.7031 & $ 1.22\times 10^{-4} $  & 2.2426\\
\hline
MUSIC  & 0.8011& 0.001  & 8.8898\\
\hline
RCNN & { 0.1887} & $\bf{2.2805\times 10^{-5}}$ &  \bf{10.4252}\\
\hline
Ensemble & \bf{0.1744} &$ 5.4\times 10^{-5} $ &10.2738\\
\hline
\end{tabular}
\end{table}

\subsection{Performance Analysis}

The overall performance of each method using LFM signal at 10dB SNR is summarized in Table.~\ref {Performance comparison}. The  table demonstrate that classification method  achieves the lowest DoA RMSE  making it the most accurate estimator among the methods evaluated under the compromise of high training time, prediction time and ESE. The RCNN, while slightly less accurate, attains the fastest prediction time, making it the strong choice for latency-sensitive deployments. The Ensemble  yields better accuracy at all range of SNRs (O7) than RCNN and achives near-identical spectral efficiency with slightly more time consumption. Beamforming time remains negligibly invariant across all methods and does not constitute a performance differentiator. Considering estimation accuracy, effective spectral efficiency, and inference latency jointly, RCNN emerges as an attractive operating point for real-time ISAC deployment, while the Ensemble model provides a competitive alternative when estimation accuracy is prioritized. 

\begin{figure*}[!t]
\centering
\includegraphics[height=5.0cm,width=5.0in]{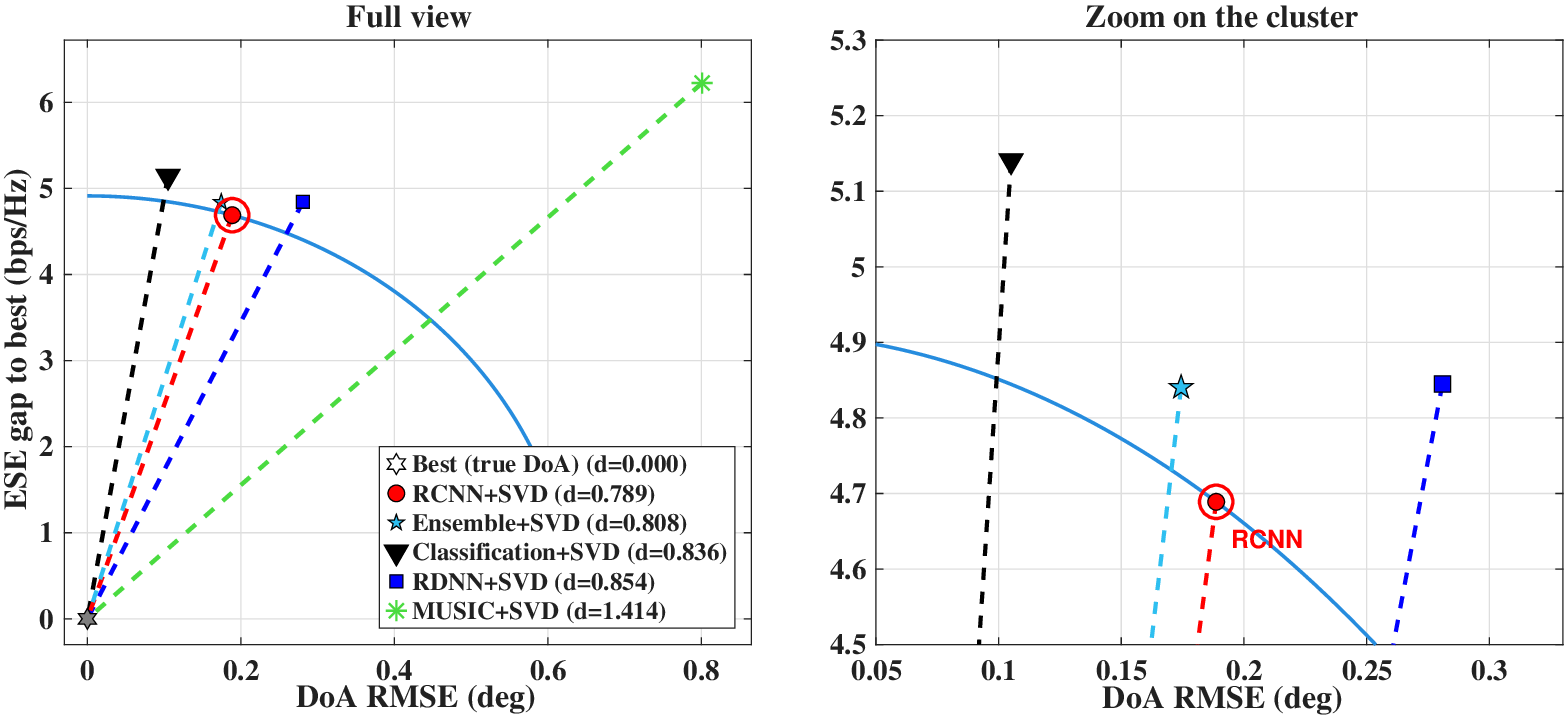}
\caption{Performance evaluation based on distance from ideal operating point (LFM signals) }
\label{ESE_rmse}
\end{figure*}

The performance evaluation of these approaches at 10dB SNR is provided in Fig~\ref {ESE_rmse}. In the plot, X - axis gives the DoA  accuracy measurement, which is RMSE in degrees. Y- axis provides ESE gap, which is the difference between the best ESE (zero RMSE, zero ESE gap relative to true-DoA beamforming) to that ESE obtained for each method. Unlike RMSE alone, the ESE gap directly quantifies the communication-performance degradation caused by DoA estimation errors and therefore provides a system-level measure of estimator quality. The arc provided shows the iso-distance contour through RCNN+SVD, and methods outside it are farther from the ideal point. Distances $d$ are Euclidean after normalizing each axis by its maximum over the plotted methods. The method with most favorable accuracy-ESE tradeoff is the one which is nearer to the ideal point which is at point (0,0). From the plot it is evident that RCNN shows less distance to ideal point. The distance to the ideal point for classification, RDNN, MUSIC, RCNN and Ensemble is 0.8362, 0.8537, 1.4142, 0.7893, 0.8075 respectively. LDNN+SVD lies farthest from the ideal point, with a distance well above that of the other methods, and is therefore omitted from the figure for clarity.

Although several neural estimators achieve competitive RMSE performance, RCNN simultaneously maintains very low inference latency which is reflected in the effective spectral efficiency comparisons. This is particularly attractive for real-time beamforming applications where frequent weight updates are required. We show that better accuracy or faster beamforming alone will not lead to the most efficient spectral utilization. Rather a balanced approach proves to be the best. 

\subsection{Design Insights for Real-Time ISAC Beamforming}

The results and discussion in the previous section lead to four design insights for real-time ISAC beamforming.

\begin{itemize}
\item \textbf{Accuracy alone is not a sufficient selection criterion.}
Lower RMSE generally improves beamforming quality, but the associated computational overhead may reduce its practical advantage in latency-sensitive ISAC applications. 

\item \textbf{Higher DoA accuracy yields diminishing returns.}
Ensemble model improve DoA accuracy  marginally, and the resulting advantage in spectral efficiency is relatively modest. More accurate DoA estimates do not  do not necessarily translate into proportionally higher communication performance.

\item \textbf{The computational bottleneck shifts to beamforming weight generation.}
Learning-based estimators make the first stage-DoA estimation stage, fast. So latency becomes dominated by the weight-generation stage in this two-stage process. Accelerating DoA estimation alone is therefore insufficient for low-latency operation in ISAC beamforming.

\item \textbf{RCNN offers a favorable accuracy--latency tradeoff.}
Among the evaluated DoA estimators, RCNN consistently lies at a favorable point of the accuracy-latency tradeoff, making it a strong candidate for real-time deployment. Although the Ensemble model achieves marginally lower RMSE, RCNN remains closer to the ideal operating point when estimation accuracy and ESE degradation are considered jointly. 

\end{itemize}

\section{Conclusion}

This paper presented a unified framework for evaluating classical and deep-learning-based DoA estimators within an ISAC beamforming pipeline. Representative estimators were integrated with SVD-based beamforming and compared under identical operating conditions in terms of estimation accuracy, inference latency, beamforming-weight generation time, effective spectral efficiency. The results show that estimator selection for practical ISAC systems should consider both estimation accuracy and computational latency rather than either metric in isolation. While learning-based approaches significantly reduce DoA estimation time compared with MUSIC, the resulting gains in effective spectral efficiency are not always proportional to the improvements in estimation accuracy. Among the evaluated methods, RCNN exhibited a favorable balance between accuracy, computational efficiency, and beamforming performance, making it a promising candidate for real-time ISAC deployment.

An important observation is that, as DoA estimation becomes faster, the computational bottleneck increasingly shifts to the SVD-based beamforming stage. Consequently, future research should investigate low-complexity beamforming approaches that reduce or eliminate the reliance on computationally intensive matrix decompositions, thereby enabling further reductions in end-to-end ISAC processing latency. 




\end{document}